\documentclass[sigconf]{acmart}
\copyrightyear{2022}
\acmYear{2022}
\setcopyright{acmcopyright}\acmConference[KDD '22]{Proceedings of the 28th ACM SIGKDD Conference on Knowledge Discovery and Data Mining}{August 14--18, 2022}{Washington, DC, USA}
\acmBooktitle{Proceedings of the 28th ACM SIGKDD Conference on Knowledge Discovery and Data Mining (KDD '22), August 14--18, 2022, Washington, DC, USA}
\acmPrice{15.00}
\acmDOI{10.1145/3534678.3539347}
\acmISBN{978-1-4503-9385-0/22/08}
\usepackage[utf8]{inputenc} 
\usepackage[T1]{fontenc}    
\usepackage{hyperref}       
\usepackage{url}            
\usepackage{booktabs}       
\usepackage{amsfonts}       
\usepackage{nicefrac}       
\usepackage{microtype}      
\usepackage{float}
\usepackage{natbib}
\usepackage{soul}
\usepackage{graphics}
\usepackage{graphicx}
\usepackage{subfigure}

\usepackage{url}
\usepackage{balance}

\usepackage{amsmath}
\usepackage[ruled,vlined,linesnumbered]{algorithm2e}
\usepackage{enumitem}
\usepackage{paralist}
\usepackage{multirow,multicol,xspace}
\usepackage{amsthm,amsmath}
\setitemize{noitemsep,topsep=3pt,parsep=3pt,partopsep=3pt}

\RequirePackage{algorithmic}

\usepackage{cleveref}
\crefname{equation}{Eq.}{Eqs.}
\crefname{table}{Table}{Tables}
\crefname{figure}{Figure}{Figures}
\crefname{section}{Section}{Sections}
\crefname{algorithm}{Algorithm}{Algorithms}

\AtBeginDocument{%
  \providecommand\BibTeX{{%
    \normalfont B\kern-0.5em{\scshape i\kern-0.25em b}\kern-0.8em\TeX}}}

\newcommand*{\Scale}[2][4]{\scalebox{#1}{$#2$}}

\begin{document}
\title{Graph Rationalization with Environment-based Augmentations} 


\author{Gang Liu} 
\affiliation{%
  \institution{{University of Notre Dame \\ Notre Dame, IN \country{USA}}}
}
\email{gliu7@nd.edu}

\author{Tong Zhao} 
\affiliation{%
    \institution{{University of Notre Dame \\ Notre Dame, IN \country{USA}}}
}
\email{tzhao2@nd.edu}

\author{Jiaxin Xu} 
\affiliation{%
    \institution{{University of Notre Dame \\ Notre Dame, IN \country{USA}}}
}
\email{jxu24@nd.edu}

\author{Tengfei Luo} 
\affiliation{%
    \institution{{University of Notre Dame \\ Notre Dame, IN \country{USA}}}
}
\email{tluo@nd.edu}

\author{Meng Jiang} 
\affiliation{%
    \institution{{University of Notre Dame \\ Notre Dame, IN \country{USA}}}
}
\email{mjiang2@nd.edu}

\newcommand{\method}{\textsc{GREA}\xspace}

\newcommand{\unets}{\textsc{U-NetsPool}\xspace}
\newcommand{\selfattn}{\textsc{SelfAttnPool}\xspace}
\newcommand{\stablegnn}{\textsc{StableGNN}\xspace}
\newcommand{\oodgnn}{\textsc{OOD-GNN}\xspace}
\newcommand{\irm}{\textsc{IRM}\xspace}
\newcommand{\dir}{\textsc{DIR}\xspace}
\newcommand{\gcn}{\textsc{GCN}\xspace}
\newcommand{\gin}{\textsc{GIN}\xspace}

\newcommand{\dirplusaug}{\textsc{DIR}$+$\textsc{RepAug}\xspace}
\newcommand{\methodnoaug}{\textsc{GREA}$-$\textsc{RepAug}\xspace}

\definecolor{amber}{rgb}{1.0, 0.49, 0.0}
\newcommand{\overbar}[1]{\mkern 1.5mu\overline{\mkern-1.5mu#1\mkern-1.5mu}\mkern 1.5mu}

\newcommand{\glassTemp}{{GlassTemp}\xspace}
\newcommand{\meltTemp}{{MeltingTemp}\xspace}
\newcommand{\density}{{PolyDensity}\xspace}
\newcommand{\oxygen}{{O$_2$Perm}\xspace}

\newcommand{\hiv}{{ogbg-HIV}\xspace}
\newcommand{\toxcast}{{ogbg-ToxCast}\xspace}
\newcommand{\toxt}{{ogbg-Tox21}\xspace}
\newcommand{\bace}{{ogbg-BACE}\xspace}
\newcommand{\bbbp}{{ogbg-BBBP}\xspace}
\newcommand{\clintox}{{ogbg-ClinTox}\xspace}
\newcommand{\sider}{{ogbg-SIDER}\xspace}

\newcommand{\regreRSquare}{{R$^2$}\xspace}
\newcommand{\regreRMSE}{{RMSE}\xspace}
\newcommand{\classifyAUC}{{AUC}\xspace}

\newcommand\tong[1]{\textcolor{cyan}{[Tong: #1]}}
\newcommand\gang[1]{\textcolor{amber}{[Gang: #1]}}
\newcommand{\jiaxin}[1]{\textcolor{magenta}{[jiaxin: #1]}}

\renewcommand{\shortauthors}{Gang Liu et al.}

\begin{abstract}
Rationale is defined as a subset of input features that best explains or supports the prediction by machine learning models. Rationale identification has improved the generalizability and interpretability of neural networks on vision and language data.
In graph applications such as molecule and polymer property prediction, identifying representative subgraph structures named as graph rationales plays an essential role in the performance of graph neural networks.
Existing graph pooling and/or distribution intervention methods suffer from the lack of examples to learn to identify optimal graph rationales. In this work, we introduce a new augmentation operation called \emph{environment replacement} that automatically creates virtual data examples to improve rationale identification. We propose an efficient framework that performs rationale-environment separation and representation learning on the real and augmented examples in \emph{latent spaces} to avoid the high complexity of explicit graph decoding and encoding.
Comparing against recent techniques, experiments on seven molecular and four polymer datasets demonstrate the effectiveness and efficiency of the proposed augmentation-based graph rationalization framework. Data and the implementation of the proposed framework are publicly available\footnote{\url{https://github.com/liugangcode/GREA}}.

\end{abstract}



\maketitle

\section{Introduction}
\label{sec:introduction}
\begin{figure*}[t]
    \centering
    \includegraphics[width=1.98\columnwidth]{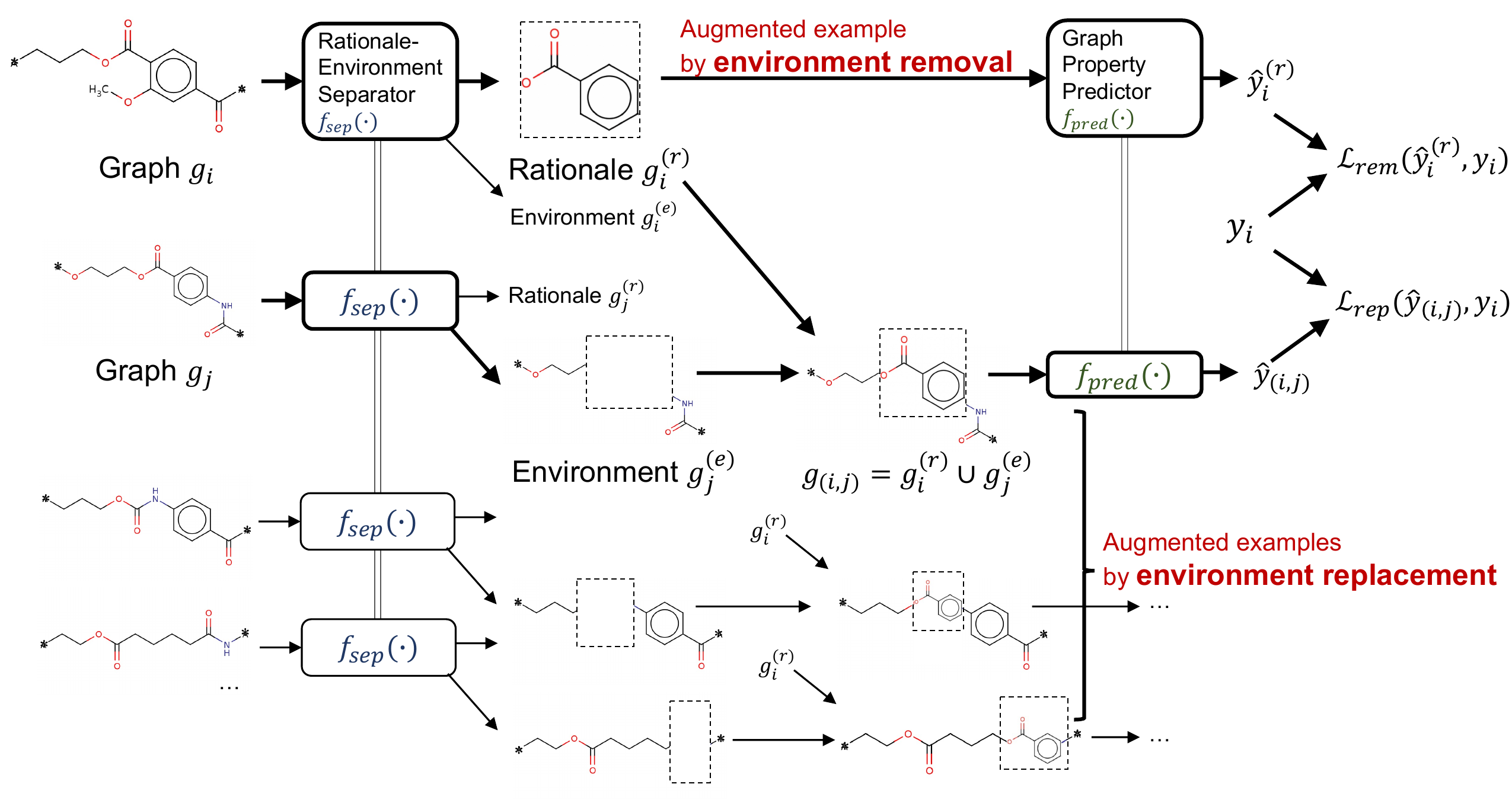}
    \vspace{-0.2in}
    \caption{Graph rationalization identifies a rationale subgraph that best explains or supports the prediction of graph property. Our work makes the first attempt to improve graph rationalization by graph data augmentations with \emph{environment subgraphs} which are the remaining parts after rationale identification. It proposes new augmentation operations, designs and develops a novel graph rationalization framework, and conducts experiments on a large set of molecule and polymer data.}
    \label{fig:idea}
    \vspace{-0.1in}
\end{figure*}

Graph property prediction has attracted attention in different research fields like chemoinformatics and bioinformatics where small molecules are represented as labelled graphs of atoms \cite{hu2020open,zhou2020data,guo2021few}.
Besides, materials informatics for \emph{polymers} has emerged in recent years from property prediction to inverse design \cite{kim2018polymer,chen2021polymer}.
Polymer are materials consisting of macromolecules, composed of many repeating units. They are ubiquitous in applications ranging from plastic cups and electronics to aerospace structures.
New engineering and environmental challenges demand that polymers possess unconventional properties such as high-temperature stability, excellent thermal conductivity, and biodegradability \cite{ma2019evaluating,wei2021thermal}.
It's important to integrate data science and machine learning into polymer informatics on the tasks of graph classification and regression.

To automate feature extraction from graph data, graph neural network (GNN) models learn node representations through nonlinear functions and layers that aggregate information from node neighborhood \cite{kipf2017semi,velivckovic2018graph,hamilton2017inductive,zhang2020deep,wu2020comprehensive}. Graph pooling is a central component of the GNN architecture that learns a cluster assignment for nodes and passes cluster nodes and their representations to the next layer \cite{ying2018hierarchical,lee2019self}. The final layer returns the representations of entire graphs. Despite the advances of various GNN models, the limitation of data size makes them easily fall into \emph{over-fitting and poor generalizability}. For example, the number of graphs in molecule benchmark datasets is usually in the range of 1,000 and 10,000; and the size of polymer datasets is even smaller (e.g., $\sim$600) \cite{ma2020pi1m}.

Rationalization techniques have been designed to solve the above problem in vision and language data, where the rationale is defined as a subset of input features that best explains or supports the prediction by machine learning models \cite{chang2020invariant,arjovsky2019invariant,rosenfeld2021risks}.
However, graph rationalization has not been extensively studied, which aims at identifying representative subgraph structures for accurate and interpretable graph property prediction.
Related work mainly focused on advancing graph pooling methods, but cluster assignment could not reflect the most essential part that led to accurate prediction \cite{mesquita2020rethinking,gao2021graph}.
A very recent technique named \dir \cite{wu2022discovering} employed two GNN modules to discover invariant graph rationales: one module separates each input graph into a rationale subgraph and an environment subgraph; the other is a graph property predictor based on the rationale subgraph. As shown at the top in Figure \ref{fig:idea}, given graph $g_i$, the separator $f_{sep}$ identifies rationale $g^{(r)}_i$, and the predictor $f_{pred}$ gives label $\hat{y}^{(r)}_i$ based on the rationale. \dir conducted interventions on training distribution to improve the invariance. Unfortunately, when the data size was small, $f_{sep}$ could hardly find good rationales, as reported in our later experiments.

\vspace{0.05in}
In this work, we make the first attempt to enhance graph rationalization by graph data augmentations. Existing augmentation methods were mainly heuristic modification of graph structure, which could not directly support the identification of graph rationales \cite{rong2019dropedge,wang2020graphcrop,wang2020nodeaug,zhao2021data}. We present two augmentation methods based on \emph{environment subgraphs} that are the remaining parts in the graph after rationale identification. First, rationales are used to train the property predictor, which can be considered as graph examples augmented by \emph{environment removal}.
Second, we replace the environment of input graph with the environment of another graph in the batch:
to generate an augmented example: this augmentation method is called \emph{environment replacement}.
The idea is that the rationale can be accurately identified and/or separated from the input graph when the augmented examples are expected to have the same label of the input graph example.

Figure \ref{fig:idea} presents the idea of generating virtual data for small datasets via data augmentations. Suppose we have rationale $g^{(r)}_i$ separated from input graph $g_i$. We use the same GNN-based separator to find environment subgraph $g^{(e)}_j$ from another graph $g_j$ in the batch.
The example augmented by environment replacement is denoted by $g_{(i,j)}=g^{(r)}_i \cup g^{(e)}_j$. The model is trained on this example to predict label $\hat{y}_{i,j}$ to be the same as $y_i$ that is the observed label of $g_i$.
We compute two losses on the augmented examples, $\mathcal{L}_{rem}$ and $\mathcal{L}_{rep}$ (``rem'' for removal and ``rep'' for replacement), and jointly optimize $f_{sep}$ and $f_{pred}$ by their combination.

The key challenge in the idea implementation is the high computational complexity of decoding for \emph{explicit graph forms} of rationales, environment subgraphs, and augmented examples, as well as encoding them for representation learning and property prediction.
Moreover, it is scientifically and technically difficult to explicitly combine rationale $g^{(r)}_i$ and environment $g^{(e)}_j$ from different graphs, as shown in the three augmented examples $g_{(i,j)}$ in Figure~\ref{fig:idea}. To address these challenges, we hypothesize that the \emph{contextualized representations of nodes} play a significant role in rationales, environment subgraphs, and augmented graphs. 
Thus, we create the representations of all these objects from \emph{one latent space}.

\begin{figure*}[t]
    \centering
    \includegraphics[width=1.99\columnwidth]{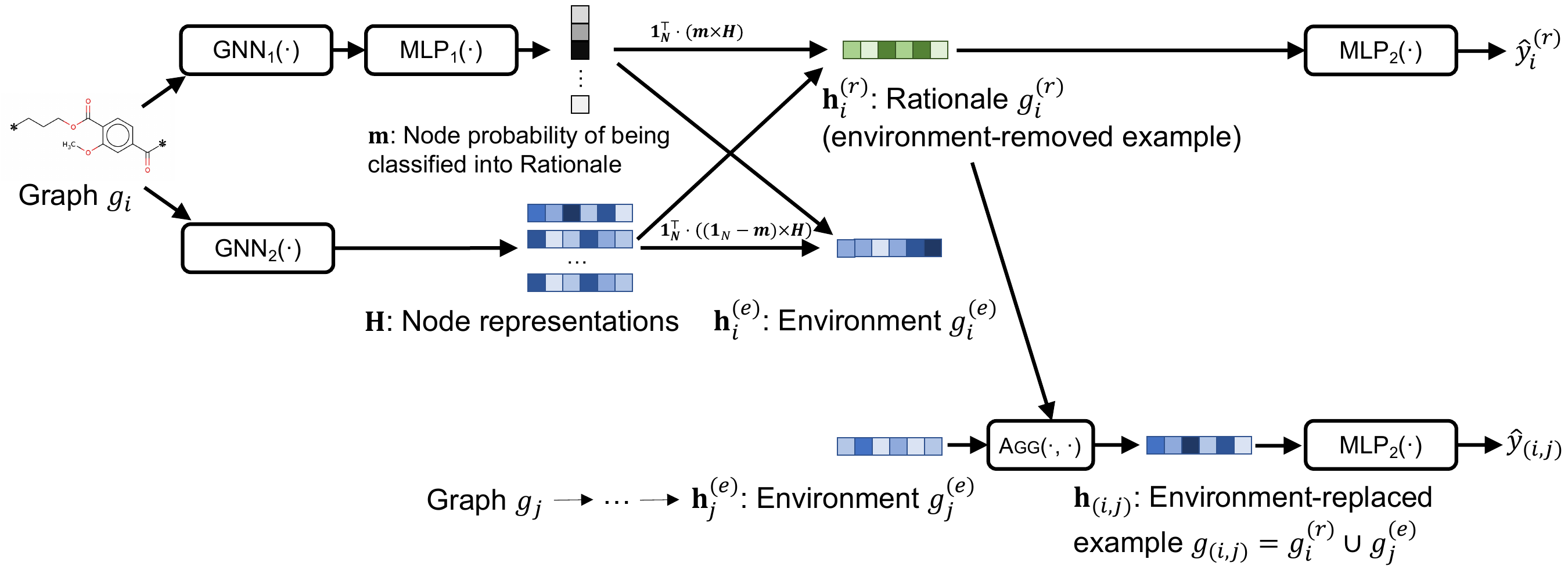}
    \vspace{-0.15in}
    \caption{The architecture of the proposed graph rationalization framework: It performs the creation and representation learning of environment-based augmented examples in a \emph{latent space}, instead of decoding every example into a graph form and running a GNN encoder on it. This design aligns graph representation spaces and avoids high computational complexity.}
    \label{fig:implementation}
    \vspace{-0.1in}
\end{figure*}

In this paper, we propose a novel, efficient framework of \underline{G}raph \underline{R}ationalization enhanced by \underline{E}nvironment-based \underline{A}ugmentations (\method).
It performs rationale-environment separation and representation learning on the real and augmented examples in  one latent space to avoid the high complexity of explicit subgraph decoding and encoding.
Figure~\ref{fig:implementation} presents the architecture of \method with a few steps.
First, it employs $\text{GNN}_1$ and $\text{MLP}_1$ models to infer the probability of nodes being classified into rationale subgraph $\mathbf{m}$.
Second, it employs $\text{GNN}_2$ to create contextualized node representations $\mathbf{H}$.
Then, it \emph{directly} creates the representation vectors of rationales, environment subgraphs and environment-replaced examples, denoted by $\mathbf{h}^{(r)}_i$, $\mathbf{h}^{(e)}_i$, and $\mathbf{h}_{(i,j)}$, respectively. Note that \dir \cite{wu2022discovering} used a GNN to generate a matrix of masks that indicate the importance of edges and then select the top-$K$ edges with the highest masks to construct the rationale. Then it had to run GNNs on all the explicit graph objects.
Instead, our \method uses $\mathbf{m}$ and $\mathbf{H}$ to compute the representation vectors of the artificial graphs.

We conduct experiments on seven molecule and four polymer datasets. Results demonstrate the advantages of \method over baselines. For example, it significantly reduces the prediction error on oxygen permeability of polymer membrane with only 595 training examples. The oxygen permeability defines how easily oxygen passes through a particular material. Accurate prediction will speed up material discovery for healthcare and energy utilization.

The main contributions of this work are summarized below:
\begin{compactitem}
    \item the first attempt to improve graph rationale identification using data augmentations, including environment replacement, for accurate and interpretable property prediction;
    \item a novel and efficient framework that performs rationale-environment separation and representation learning on real and augmented examples in one latent space;
    \item extensive experiments on more than ten molecule and polymer datasets to demonstrate the effectiveness and efficiency of the proposed framework.
\end{compactitem}

\section{Related Work}
\label{sec:related}

There are four research topics related to the proposed work. We briefly present their recent studies and compare with ours.

\subsection{Graph Property Prediction}

Learning representations and predicting properties of entire graphs is important for chemistry, biology, and material sciences, where molecule and polymer data can be structured as graphs \cite{hu2020open}.
When RDKit is widely used to generate molecular fingerprints \cite{landrum2013rdkit}, 
graph neural networks (GNNs) such as Graph Convolutional Network (GCN)~\cite{kipf2017semi}, Graph Attention Networks (GAT)~\cite{velivckovic2018graph}, and \textsc{GraphSAGE}~\cite{hamilton2017inductive} have automated representation learning with nonlinear functions from graph data~\cite{zhang2020deep,wu2020comprehensive,ma2021unified,wang2020calendar,wang2021dynamic,wang2021modeling,wang2021modeling_tkde,zhao2021synergistic,jiang2022federated}.

In the GNN models, graph pooling is a central component of their architectures as a cluster assignment function to find local patches in graphs \cite{mesquita2020rethinking}.
For example, \textsc{DiffPool} presented a differentiable graph pooling module that learned a differentiable soft cluster assignment for nodes at each layer of a deep GNN, mapped nodes to a set of clusters, and then formed the coarsened input for the next GNN layer \cite{ying2018hierarchical}. Lee et al. proposed self-attention graph convolution that allows graph pooling to consider both node features and graph topology \cite{lee2019self}. Gao and Ji proposed graph pooling and unpooling operations in Graph \textsc{U-Nets} \cite{gao2021graph}. Xu et al. presented a theoretical framework for analyzing the representational power of GNNs through the graph pooling functions \cite{xu2018how}. While graph pooling identifies soft clusters that effectively aggregate information from nodes \cite{ying2019gnnexplainer}, our work identifies representative subgraph structures for accurate and interpretable predictions of GNN models.

\subsection{Graph Rationalization}

Most rationalization techniques identify the small subset of input features by maximizing the predictive performance based only on the subset itself, called rationale. To rule out spurious correlation between the input features and the output, Chang et al. proposed the concept of invariant rationalization by modeling different environments as non-causal input to train predictors \cite{chang2020invariant}. Rosefeld et al. offered formal guarantees for improvement of the invariant causal prediction on out-of-distribution generalization \cite{arjovsky2019invariant, rosenfeld2021risks}.

By introducing causal modeling into GNN optimization, Fan et al. presented a causal representation framework for GNN models to perform on out-of-distribution graphs \cite{fan2021generalizing}. Li et al. proposed \oodgnn that employed a novel nonlinear graph representation decorrelation method that used random Fourier features to encourage GNNs to eliminate the statistical dependence between relevant and irrelevant graph representations \cite{li2021ood}.
Very recently, Wu et al. proposed the first work called \dir to approach causal rationales for GNNs to improve the interpretability and predictive performance on out-of-distribution data \cite{wu2022discovering}. \dir conducted interventions on the training distribution to create multiple distributions. Unfortunately, distribution intervention might not be the optimal solution to graph rationale identification. Also, the edge selection method suffers from high computational complexity for rationale creation. Moreover, the studies were mainly performed on synthetic data.
In this paper, we make the first attempt to define ``environment'' in graph data, augment data examples by environment replacement, develop an efficient framework, and conduct experiments on a large set of real molecule and polymer data. We find that augmentation-enhanced graph rationalization is more effective than \dir.

\subsection{Graph Data Augmentation}

Graph data augmentation (GDA) techniques~\cite{zhao2022graph,zhao2021counterfactual,zhao2021action,chen2020measuring} have improved the performance on semi-supervised node classification, such as \textsc{DropEdge} \cite{rong2019dropedge}, \textsc{NodeAug} \cite{wang2020nodeaug}, and \textsc{GAug} \cite{zhao2021data}.
Besides, many GDA techniques have been designed for graph-level tasks, aiming at creating new training examples by modifying input graph data examples. For example, \textsc{GraphCrop} regularized GNN models for better generalization by cropping subgraphs or motifs to simulate real-world noise of sub-structure omission \cite{wang2020graphcrop}.
\textsc{M-Evolve} presented two heuristic algorithms including random mapping and motif-similarity mapping to generate weakly labeled data for small datasets \cite{zhou2020data}.
\textsc{MH-Aug} adopted the Metropolis-Hastings algorithm to create augmented graphs from an explicit target distribution for semi-supervised learning \cite{park2021metropolis}.
Meanwhile, graph contrastive learning learned unsupervised representations of graphs using graph data augmentations to incorporate various priors~\cite{you2020graph}. \citet{zhu2021graph} proposed adaptive augmentation that incorporated various priors for topological and semantic aspects of graphs. Specifically, it designed augmentation schemes based on node centrality measures to highlight important connective structures and corrupted node features by adding noise to unimportant node features. A comprehensive survey of GDA is given by \citet{zhao2022graph}.

\subsection{Graph Learning on Polymer Data}

Material informatics uses machine learning approaches to fast screen material candidates or generate new materials meeting certain criteria, so as to reduce the time of material development. When most related research performed on molecule data \cite{guo2021few}, polymer researchers have developed a benchmark database and developed machine learning techniques for polymer data, called polymer embeddings \cite{kim2018polymer,chen2021polymer}. They can be used to perform several polymer informatics regression tasks for density, glass transition temperature, melting temperature, and dielectric constants \cite{ma2019evaluating,ma2020pi1m,wei2021thermal}.

\section{Problem Definition}
\label{sec:problem}
\paragraph{Graph Property Prediction}
Let $g = (\mathcal{V}, \mathcal{E})$ be a graph of $N$ nodes and $M$ edges, where $\mathcal{V}$ is the set of nodes (e.g., atoms) and $\mathcal{E} \subseteq \mathcal{V} \times \mathcal{V}$ is the set of edges (e.g., bonds between atoms). We use $y \in \mathcal{Y}$ to denote the graph-level property of $g$, where $\mathcal{Y}$ is the value space. It can have a categorical or numerical value, corresponding to the task of classification or regression, respectively.

A graph property predictor $f_{pred}$ takes a graph $g$ as input and predicts its label $\hat{y}$. Specifically, a GNN-based predictor employs a GNN encoder to generate node representations $\mathbf{H}$ from $g$:
\begin{equation}
\label{eq:gnn}
    \mathbf{H} = \begin{bmatrix}
    \cdots, \vec{h}_v, \cdots
    \end{bmatrix}_{v \in \mathcal{V}}^\top
    = \text{GNN}(g) \in \mathbb{R}^{N \times d},
\end{equation}
where $\vec{h}_v \in \mathbb{R}^d$ is the representation vector of node $v$ in graph $g$. GNN encoder $\text{GNN}(\cdot)$ can be chosen as \gcn \cite{kipf2017semi} or \gin \cite{xu2018how}.

Once the node representations are ready, a multilayer perceptron (MLP) can project them into a one-dimensional space to obtain a scalar for each node as $m_v = \text{MLP}(\vec{h}_v).$
As we are more interested in graph-level classification or regression, we first use a readout operator (e.g., average pooling) to get the graph representation $\mathbf{h}$ and then apply a MLP to project it to a graph label:
\begin{equation}
\label{eq:readout_and_mlp_g}
    \mathbf{h} = \text{READOUT}(\mathbf{H}) \in \mathbb{R}^{d}, \quad \hat{y} = \text{MLP}(\mathbf{h}) \in \mathcal{Y}.
\end{equation}

\paragraph{Graph Rationalization}
Following the existing literature on graph rationalization~\cite{ying2018hierarchical,lee2019self,gao2021graph,fan2021generalizing,wu2022discovering} and GNN explanation~\cite{ying2019gnnexplainer}, we use rationale $g^{(r)} = (\mathcal{V}^{(r)}, \mathcal{E}^{(r)})$ to indicate the causal subgraph of the property $y$, where $g^{(r)}$ is a subgraph of $g$ such that $\mathcal{V}^{(r)} \subseteq \mathcal{V}$ and $\mathcal{E}^{(r)} \subseteq \mathcal{E}$. We use $g^{(e)}$ to denote the environment subgraph, which is the complementary subgraph of $g^{(r)}$ in $g$.
In contrast with the rationale subgraph $g^{(r)}$, the environment subgraph $g^{(e)}$ corresponds to the non-causal part of the graph data, which has no causal relationship with the target graph property~\cite{chang2020invariant,wu2022discovering}.

Let $f_{sep}$ be a GNN-based graph rationalization model that splits an input graph $g$ into a rationale subgraph $g^{(r)}$ and an environment subgraph $g^{(e)}$. Existing graph rationalization methods used only the rationale subgraph as input for property prediction~\cite{lee2019self,ying2018hierarchical,gao2021graph,wu2022discovering}:
\begin{equation}
\label{eq:f_pred}
    \hat{y} = \hat{y}^{(r)} = f_{pred} \big(g^{(r)}\big),
\end{equation}
where $f_{pred}(\cdot) = \text{MLP}(\text{READOUT}(\text{GNN}(\cdot)))$ and $\hat{y}^{(r)}$ denotes the predicted property of the rationale subgraph $g^{(r)}$.

Unfortunately, when suffering from lack of training examples, these methods chose to discard environment subgraphs at the training stage. 
In the next section, we present a novel framework showing our idea that environment subgraphs can provide natural noise through data augmentation to improve graph rationalization.

\section{Proposed Framework}
\label{sec:method}
In this section, we introduce a novel graph rationalization framework \method. The key idea is to augment the rationale subgraph by removing its own environment subgraph and/or combining it with different environment subgraphs. \cref{fig:implementation} shows the overall architecture of \method:
$\text{GNN}_1$ and $\text{MLP}_1$ first separate input graph $g$ into rationale subgraph $g^{(r)}$ and environment subgraph $g^{(e)}$;
$\text{GNN}_2$ next generates node representations $\mathbf{H}$ using Eq.(\ref{eq:gnn}); 
the rationale subgraph's representation $\mathbf{h}^{(r)}_i$ is then combined with different environment subgraph's representations $\mathbf{h}^{(e)}_j$ for the augmented graph's representations $\mathbf{h}_{(i,j)}$; 
finally, both $\mathbf{h}^{(r)}_i$ and $\mathbf{h}_{(i,j)}$ are fed into $\text{MLP}_2$ for the prediction of $y_i$ during training as Eq.(\ref{eq:readout_and_mlp_g}).

\subsection{Rationale-Environment Separation} 
\label{sec:gnnsep}
To separate input graph $g$ into rationale subgraph $g^{(r)}$ and environment subgraph $g^{(e)}$, the rationale-environment separator consists of two components: a GNN encoder ($\text{GNN}_1$) that generates latent node representations and a MLP decoder ($\text{MLP}_1$) that maps the node representations to a mask vector $\mathbf{m} \in {(0, 1)}^{N}$ on the nodes in the set $\mathcal{V}$. $m_v = Pr(v \in \mathcal{V}^{(r)})$ is the node-level mask that indicates the probability of node $v \in \mathcal{V}$ being classified into the rationale subgraph.
The mask can be on either a node or an edge~\cite{wu2022discovering}. we choose to learn masks on the nodes to avoid the computational complexity of edge selection. Hence, $\mathbf{m}$ can be calculated as
\begin{equation}
\label{eq:separator_enc_dec}
    \mathbf{m} = \sigma(\text{MLP}_1(\text{GNN}_1(g))),
\end{equation}
where $\sigma$ denotes the sigmoid function. Based on $\mathbf{m}$, we have $(\mathbf{1}_{N} - \mathbf{m})$ that indicates the probability of nodes being classified into the environment subgraph. $\text{GNN}_1$ and $\text{MLP}_1$ make up the GNN-based graph rationalization model $f_{sep}$ mentioned in \cref{sec:problem}.

\method uses another GNN encoder to generate contextualized node representations $\mathbf{H}$: $\mathbf{H} = \text{GNN}_2(g)$.
With $\mathbf{m}$ and $\mathbf{H}$, the rationale subgraph and environment subgraph can be easily separated in the \emph{latent space}. Using sum pooling, we have 
\begin{align}
    \label{eq:h_r}
    \mathbf{h}^{(r)} = \mathbf{1}^{\top}_N \cdot (\mathbf{m} \times \mathbf{H}), \quad
    \mathbf{h}^{(e)} = \mathbf{1}^{\top}_N \cdot ((\mathbf{1}_N - \mathbf{m}) \times \mathbf{H}),
\end{align}
where $\mathbf{1}_N$ denotes the $N$-size column vector with all entries as 1, and $\mathbf{h}^{(r)},\mathbf{h}^{(e)} \in \mathbb{R}^{d}$ are the representation vectors of graph $g^{(r)}$ and $g^{(e)}$, respectively.

\subsection{Environment-based Augmentations}
\label{sec:env_aug}

Suppose $g_1, g_2, \dots, g_B$ are the input graphs in one batch for training, where $B$ is known as batch size.
The rationale-environment separator has generated the graph representations of rationale and environment subgraphs for each graph $g_i$. That is, we have
$\{(\mathbf{h}_1^{(r)}, \mathbf{h}_1^{(e)}), (\mathbf{h}_2^{(r)}, \mathbf{h}_2^{(e)}), \dots, (\mathbf{h}_B^{(r)}, \mathbf{h}_B^{(e)})\}$. We design environment-based augmentations in the latent space of graph representations.

\subsubsection{Environment Removal Augmentation} 
As graph rationalization aims to find the rationale subgraph which is regarded as the causal factor of graph property, the rationale itself should be good for property prediction. As in the graph pooling methods~\cite{lee2019self,gao2021graph} and the graph rationalization as defined in \cref{eq:f_pred}, the environment removal augmentation uses the rationale subgraph only for training the graph property predictor. That is, given the rationale subgraph representation $\mathbf{h}_i^{(r)}$ of graph $g_i$, the predicted label is
\begin{equation}
\label{eq:mlp2_rem}
\hat{y}_{i}^{(r)} = \text{MLP}_2 \big(\mathbf{h}_i^{(r)}\big).
\end{equation}

\subsubsection{Environment Replacement Augmentation}
As aforementioned in \cref{sec:problem}, the environment subgraphs can be viewed as natural noises on the rationale subgraphs. Hence, in order to enhance the model's robustness against the noise signal brought by the environment subgraphs, for each graph $g_i$, we combine its rationale subgraph $g_i^{(r)}$ not only with its own environment subgraph $g_i^{(e)}$, but also with all other environment subgraphs $g_j^{(e)}, j \in \{1, 2, \dots, B\} \setminus \{i\}$ in the batch. By replacing the environment subgraph with other environment subgraphs in the batch, the environment replacement augmentation generates $B-1$ augmented data samples for each graph during training. As the environment replacement happens on the latent space, an aggregation function $\text{AGG}(\cdot, \cdot)$ is used to combine the rationale subgraph representation $\mathbf{h}_i^{(r)}$ and environment subgraph representation $\mathbf{h}_j^{(e)}$. The aggregation function can be any combining/pooling functions such as concatenation, sum pooling, and max pooling. Taking the element-wise sum pooling as an example, the graph representation $\mathbf{h}_{(i,j)}$ of a combined graph of rationale subgraph $g_i^{(r)}$ and environment subgraph $g_j^{(e)}$ can be calculated as below:
\begin{equation}
\label{eq:rat_env_agg}
    \mathbf{h}_{(i,j)} = \text{AGG}\big(\mathbf{h}_i^{(r)}, \mathbf{h}_j^{(e)}\big) = \mathbf{h}_i^{(r)} + \mathbf{h}_j^{(e)}.
\end{equation}

For the graph representations $\mathbf{h}_{(i,j)}$ generated by the environment replacement augmentation, the MLP property predictor is trained to predict $y_i$. That is,
\begin{equation}
\label{eq:mlp2_rep}
    \hat{y}_{(i,j)} = \operatorname{MLP}_2 \big(\mathbf{h}_{(i,j)} \big).
\end{equation}

\begin{table}[t]
\caption{Statistics of eleven datasets for graph property prediction: The four top rows are polymer datasets. The prediction tasks are graph regression. The seven bottom rows are molecule datasets. Their tasks are graph classification.}
\label{tab:dataset_stat}
\vspace{-0.1in}
\centering
\Scale[0.95]{\begin{tabular}{lrrr}
\toprule
Dataset & \# Graphs & Avg./Max \# Nodes & Avg./Max \# Edges \\
\midrule
\glassTemp & 7,174 & 36.7 / 166 & 79.3 / 362 \\
\meltTemp & 3,651 & 26.9 / 102 & 55.4 / 212 \\
\density & 1,694 & 27.3 / 93 & 57.6 / 210 \\
\oxygen & 595 & 37.3 / 103 & 82.1 / 234 \\
\midrule
\hiv & 41,127 & 25.5 / 222 & 54.9 / 502 \\
\toxcast & 8,576 & 18.8 / 124 & 38.5 / 268 \\
\toxt & 7,831 & 18.6 / 132 & 38.6 / 290 \\
\bbbp & 2,039 & 24.1 / 132 & 51.9 / 290 \\
\bace & 1,513 & 34.1 / 97 & 73.7 / 202 \\
\clintox & 1,477 & 26.2 / 136 & 55.8 / 286 \\
\sider & 1,427 & 33.6 / 492 & 70.7 / 1010 \\
\bottomrule
\end{tabular}}
\end{table}

The graph representations generated by both environment removal augmentation and environment replacement augmentation (i.e., $\mathbf{h}_i^{(r)}$ and $\mathbf{h}_{(i,j)}$) are fed into the same property predictor $\text{MLP}_2$. The GNN-based property predictor $f_{pred}$ defined in \cref{sec:problem} includes $\text{MLP}_2$ and $\text{GNN}_2$ that generates the contextualized node representation $\mathbf{H}$.

\subsubsection{Optimization}
\label{sec:training}
During training, the type of loss function on the observed graph property ($y_i$) and predicted labels ($\hat{y}_{i}^{(r)}$ and $\hat{y}_{(i,j)}$) depends on the type of the property label. For example, when the graph property $y$ has binary values in the binary classification task, we use the standard binary cross-entropy loss. When the graph property $y$ has real values in the graph regression task, we use the mean squared error (MSE) loss. Without loss of generality, suppose we focus on the binary classification task. Given a batch of $B$ graphs $g_1, g_2, \dots, g_B$, the loss functions for each graph example $g_i$ and its label $y_i$ are defined as
\begin{align}
    \label{eq:loss_rem}
    &\mathcal{L}_{rem} = y_{i} \cdot \log \hat{y}_{i}^{(r)} + \left(1-y_{i}\right) \cdot \log \big(1-\hat{y}_{i}^{(r)}\big), \\
    \label{eq:loss_rep}
    &\mathcal{L}_{rep} = \frac{1}{B} \sum_{j=1}^B \big( y_{i} \cdot \log \hat{y}_{(i,j)} + \left(1-y_{i}\right) \cdot \log (1-\hat{y}_{(i,j)}) \big),
\end{align}
where $\mathcal{L}_{rem}$ is the loss for the examples created by environment removal augmentation, and $\mathcal{L}_{rep}$ is the loss for the examples created by the environment replacement augmentation. 

Moreover, the following regularization term is used to control the size of the selected rationale subgraph:
\begin{equation}
\label{eq:loss_reg}
    \mathcal{L}_{reg} = \Big|\frac{\mathbf{1}^\top_N \cdot \mathbf{m}}{N} - \gamma \Big|,
\end{equation}
where $\gamma \in [0,1]$ is a hyperparamter to control the expected size of the rationale subgraph $g^{(r)}$. We penalize the number of nodes in the rationale when it deviates from our expectations.

\begin{table*}[t]
\caption{Results on polymer property prediction: \method consistently achieves the highest \regreRSquare and smallest \regreRMSE.}
\label{tab:result_plym}
\vspace{-0.1in}
\centering
\Scale[0.95]{\begin{tabular}{l|l|rr|rr|rr|rr}
\toprule
\multicolumn{2}{l|}{} & \multicolumn{2}{c|}{\glassTemp} & \multicolumn{2}{c|}{\meltTemp}& \multicolumn{2}{c|}{\density} & \multicolumn{2}{c}{\oxygen} \\
\multicolumn{2}{l|}{} & \multicolumn{1}{c}{\regreRSquare $\uparrow$} & \multicolumn{1}{c|}{\regreRMSE $\downarrow$} & \multicolumn{1}{c}{\regreRSquare $\uparrow$} & \multicolumn{1}{c|}{\regreRMSE $\downarrow$} & \multicolumn{1}{c}{\regreRSquare $\uparrow$} & \multicolumn{1}{c|}{\regreRMSE $\downarrow$} & \multicolumn{1}{c}{\regreRSquare $\uparrow$} & \multicolumn{1}{c}{\regreRMSE $\downarrow$} \\
\midrule
\parbox[t]{2mm}{\multirow{9}{*}{\rotatebox[origin=c]{90}{\gcn~\cite{kipf2017semi} as encoder}}} & \unets~\cite{gao2021graph} & 0.839$\pm$0.005 & 44.9$\pm$0.7 & 0.685$\pm$0.012 & 63.4$\pm$1.2 & 0.615$\pm$0.053 & 0.100$\pm$0.007 & 0.833$\pm$0.084 & 865$\pm$214 \\
& \selfattn~\cite{lee2019self} & 0.848$\pm$0.007 & 43.5$\pm$1.0 & 0.709$\pm$0.008 & 61.0$\pm$0.9 &  0.688$\pm$0.019 & \underline{0.090}$\pm$0.003 & 0.656$\pm$0.135 & 1251$\pm$266 \\
& \stablegnn~\cite{fan2021generalizing} & 0.809$\pm$0.013 & 48.8$\pm$1.6 & 0.635$\pm$0.033 & 70.0$\pm$4.5 & 0.667$\pm$0.070 & 0.093$\pm$0.009 & 0.676$\pm$0.127 & 1219$\pm$241 \\
& \oodgnn~\cite{li2021ood} & \underline{0.852}$\pm$0.006 & \underline{43.0}$\pm$0.9 & \underline{0.714}$\pm$0.025 & \underline{60.4}$\pm$2.6 & 0.676$\pm$0.010 & 0.092$\pm$0.001 & \underline{0.921}$\pm$0.059 & \underline{576}$\pm$212 \\
& \irm~\cite{arjovsky2019invariant} & 0.830$\pm$0.008 & 46.1$\pm$1.1 & 0.677$\pm$0.006 & 64.2$\pm$0.6 & \underline{0.690}$\pm$0.016 & \underline{0.090}$\pm$0.002 & 0.871$\pm$0.043 & 770$\pm$141 \\
& \dir~\cite{wu2022discovering} & 0.697$\pm$0.061 & 61.2$\pm$6.0 & 0.380$\pm$0.214 & 87.8$\pm$14. & 0.656$\pm$0.036 & 0.094$\pm$0.005 & 0.135$\pm$0.068 & 2028$\pm$80 \\
& \dirplusaug & 0.800$\pm$0.006 & 56.5$\pm$3.2 & 0.520$\pm$0.101 & 77.8$\pm$8.2 & 0.671$\pm$0.033 & 0.092$\pm$0.005 & 0.915$\pm$0.031 & 626$\pm$115 \\
& \methodnoaug & 0.685$\pm$0.172 & 60.6$\pm$16.5 & 0.679$\pm$0.034 & 64.0$\pm$3.3 & 0.686$\pm$0.007 & \underline{0.090}$\pm$0.001 & 0.459$\pm$0.254 & 1556$\pm$395 \\
& \method (ours) & \textbf{0.855}$\pm$0.003 & \textbf{42.6}$\pm$0.5 & \textbf{0.716}$\pm$0.016  & \textbf{60.2}$\pm$1.6 &  \textbf{0.717}$\pm$0.023 & \textbf{0.086}$\pm$0.003 & \textbf{0.941}$\pm$0.018 & \textbf{524}$\pm$91 \\
\midrule
\parbox[t]{2mm}{\multirow{9}{*}{\rotatebox[origin=c]{90}{\gin~\cite{xu2018how} as encoder}}} & \unets~\cite{gao2021graph} & 0.852$\pm$0.006 & 42.9$\pm$0.9 & 0.703$\pm$0.009 & 61.6$\pm$0.9 & 0.635$\pm$0.029 & 0.097$\pm$0.004 & 0.868$\pm$0.085 & 753$\pm$250 \\
& \selfattn~\cite{lee2019self} & 0.848$\pm$0.003 & 43.5$\pm$0.4 & \underline{0.726}$\pm$0.009 & \underline{59.2}$\pm$1.0 & 0.654$\pm$0.024 & 0.095$\pm$0.003 & 0.601$\pm$0.267 & 1265$\pm$546 \\
& \stablegnn~\cite{fan2021generalizing} & 0.794$\pm$0.007 & 50.8$\pm$0.9 & 0.535$\pm$0.061 & 76.9$\pm$5.0 & 0.642$\pm$0.045 & 0.096$\pm$0.006 & 0.501$\pm$0.266 & 1487$\pm$404 \\
& \oodgnn~\cite{li2021ood} & \underline{0.862}$\pm$0.007 & \underline{41.6}$\pm$1.1 & 0.721$\pm$0.006 & 59.7$\pm$0.6 & 0.666$\pm$0.025 & 0.093$\pm$0.003 & \underline{0.917}$\pm$0.029 & \underline{620}$\pm$109 \\
& \irm~\cite{arjovsky2019invariant} & 0.842$\pm$0.004 & 44.5$\pm$0.5 & 0.681$\pm$0.008 & 63.8$\pm$0.8 & 0.682$\pm$0.031 & 0.091$\pm$0.004 & 0.890$\pm$0.042 & 709$\pm$146 \\
& \dir~\cite{wu2022discovering} & 0.594$\pm$0.070 & 71.0$\pm$6.0 & 0.287$\pm$0.121 & 95.1$\pm$7.9 & 0.617$\pm$0.045 & 0.099$\pm$0.006 & 0.501$\pm$0.309 & 1446$\pm$537 \\
& \dirplusaug & 0.744$\pm$0.029 & 56.4$\pm$3.2 & 0.542$\pm$0.083 & 76.2$\pm$7.0 & 0.647$\pm$0.058 & 0.095$\pm$0.008 & 0.743$\pm$0.150 & 1054$\pm$338 \\
& \methodnoaug & 0.494$\pm$0.110 & 79.0$\pm$9.3 & 0.660$\pm$0.107 & 65.2$\pm$9.5 & \underline{0.717}$\pm$0.022 & \underline{0.086}$\pm$0.003 & 0.400$\pm$0.286 & 1623$\pm$474 \\
& \method (ours) & \textbf{0.864}$\pm$0.005 & \textbf{41.2}$\pm$0.8 & \textbf{0.736}$\pm$0.012 & \textbf{58.0}$\pm$1.2 & \textbf{0.723}$\pm$0.030 & \textbf{0.085}$\pm$0.005 & \textbf{0.930}$\pm$0.020 & \textbf{569}$\pm$86 \\
\bottomrule
\end{tabular}}
\end{table*}

\begin{table*}[t]
\caption{Results on molecule property prediction: \method consistently achieves the highest \classifyAUC ($\uparrow$).}
\label{tab:res_ogbg_mol}
\vspace{-0.1in}
\centering
\Scale[0.93]{\begin{tabular}{llccccccc}
\toprule
\multicolumn{2}{l}{} & \hiv & \toxcast & \toxt & \bbbp & \bace & \clintox & \sider \\ 
\midrule
\parbox[t]{2mm}{\multirow{9}{*}{\rotatebox[origin=c]{90}{\gcn~\cite{kipf2017semi} as encoder}}} & \unets~\cite{gao2021graph}
& 0.7527$\pm$0.0104 & 0.6507$\pm$0.0086 & 0.7492$\pm$0.0093       
& 0.6709$\pm$0.0176 & 0.7757$\pm$0.0173 & 0.8450$\pm$0.0403
& 0.6181$\pm$0.0121 \\
& \selfattn~\cite{lee2019self}
& \underline{0.7733}$\pm$0.0187 & 0.6510$\pm$0.0076 & 0.7563$\pm$0.0080       
& 0.6602$\pm$0.0220 & 0.7383$\pm$0.0541 & 0.8291$\pm$0.0791
& 0.5718$\pm$0.0219 \\
& \stablegnn~\cite{fan2021generalizing}
& 0.7218$\pm$0.0099 & 0.6520$\pm$0.0109 & 0.7454$\pm$0.0059       
& 0.6552$\pm$0.0184 & 0.6607$\pm$0.0500 & 0.7681$\pm$0.0778
& 0.5644$\pm$0.0274 \\
& \oodgnn~\cite{li2021ood}
& 0.7580$\pm$0.0176 & 0.6613$\pm$0.0046 & 0.7673$\pm$0.0109       
& 0.6795$\pm$0.0165 & \underline{0.8096}$\pm$0.0132 & \underline{0.8874}$\pm$0.0143
& 0.6133$\pm$0.0095 \\
& \irm~\cite{arjovsky2019invariant}
& 0.7702$\pm$0.0107 & 0.6599$\pm$0.0063 & 0.7654$\pm$0.0072       
& \underline{0.6892}$\pm$0.0053 & 0.7947$\pm$0.0186 & 0.8819$\pm$0.0231
& 0.6035$\pm$0.0195 \\
& \dir~\cite{wu2022discovering}
& 0.7466$\pm$0.0093 & 0.5954$\pm$0.0154 & 0.4727$\pm$0.0129        
& 0.6559$\pm$0.0298 & 0.6751$\pm$0.0323 & 0.6251$\pm$0.0956
& 0.5331$\pm$0.0216 \\
& \dirplusaug & 0.7494$\pm$0.0225 & \underline{0.6632}$\pm$0.0098 & 0.7437$\pm$0.0054 & 0.6630$\pm$0.0118 & 0.7677$\pm$0.0226 & 0.8606$\pm$0.0144 & 0.5934$\pm$0.0170 \\ 
& \methodnoaug & 0.7377$\pm$0.0210 & 0.6614$\pm$0.0048 & \underline{0.7808}$\pm$0.0061 & 0.6736$\pm$0.0077 & 0.7655$\pm$0.0529 & 0.8708$\pm$0.0514 & \underline{0.6222}$\pm$0.0166  \\
& \method (ours)
& \textbf{0.7794}$\pm$0.0065 & \textbf{0.6662}$\pm$0.0041 & \textbf{0.7822}$\pm$0.0093
& \textbf{0.6986}$\pm$0.0175 & \textbf{0.8191}$\pm$0.0240 & \textbf{0.8961}$\pm$0.0150
& \textbf{0.6316}$\pm$0.0151 \\
\midrule
\parbox[t]{2mm}{\multirow{9}{*}{\rotatebox[origin=c]{90}{\gin~\cite{xu2018how} as encoder}}} & \unets~\cite{gao2021graph}
& 0.7375$\pm$0.0362 & 0.6524$\pm$0.0126 & 0.7560$\pm$0.0093
& 0.6809$\pm$0.0163 & \underline{0.8026}$\pm$0.0105 & 0.8146$\pm$0.0703
& 0.5929$\pm$0.0114 \\
& \selfattn~\cite{lee2019self}
& 0.7533$\pm$0.0247 & 0.6351$\pm$0.0137 & 0.7507$\pm$0.0110
& 0.6624$\pm$0.0167 & 0.7348$\pm$0.0194 & 0.7912$\pm$0.0995
& 0.5702$\pm$0.0137 \\
& \stablegnn~\cite{fan2021generalizing}
& 0.7218$\pm$0.0078 & 0.6485$\pm$0.0025 & 0.7381$\pm$0.0123       
& 0.6695$\pm$0.0120 & 0.7229$\pm$0.0122 & 0.8559$\pm$0.0224
& 0.5593$\pm$0.0172 \\
& \oodgnn~\cite{li2021ood}
& 0.7799$\pm$0.0078 & \underline{0.6697}$\pm$0.0051 & 0.7646$\pm$0.0038       
& 0.6710$\pm$0.0188 & 0.7800$\pm$0.0228 & 0.8416$\pm$0.0496
& 0.5916$\pm$0.0169 \\
& \irm~\cite{arjovsky2019invariant}
& \underline{0.7817}$\pm$0.0120 & 0.6641$\pm$0.0065 & 0.7542$\pm$0.0084
& \underline{0.6835}$\pm$0.0071 & 0.7977$\pm$0.0208 & 0.8485$\pm$0.0215
& 0.5778$\pm$0.0206 \\
& \dir~\cite{wu2022discovering}
& 0.7533$\pm$0.0117 & 0.5927$\pm$0.0097 & 0.5078$\pm$0.0313
& 0.5843$\pm$0.0443 & 0.6115$\pm$0.0587 & 0.6911$\pm$0.0810
& 0.5406$\pm$0.0127 \\
& \dirplusaug & 0.7725$\pm$0.0249  & 0.6454$\pm$0.0061 & 0.7453$\pm$0.0080 & 0.6813$\pm$0.0203 & 0.7590$\pm$0.0642 & 0.8561$\pm$0.0159  & 0.5730$\pm$0.0115 \\ 
& \methodnoaug & 0.7770$\pm$0.0178 & 0.6681$\pm$0.0066 & \underline{0.7690}$\pm$0.0117 & 0.6737$\pm$0.0235 & 0.7997$\pm$0.0380 & \underline{0.8574}$\pm$0.0442 & \underline{0.5988}$\pm$0.0169 \\
& \method (ours)
& \textbf{0.7932}$\pm$0.0092 & \textbf{0.6750}$\pm$0.0067 & \textbf{0.7723}$\pm$0.0119
& \textbf{0.6970}$\pm$0.0128 & \textbf{0.8237}$\pm$0.0237 & \textbf{0.8789}$\pm$0.0368
& \textbf{0.6014}$\pm$0.0204 \\
\bottomrule
\end{tabular}}
\end{table*}

We use the alternate training schema in~\citet{chang2020invariant} to train \method. That is, we iteratively train $f_{sep}$ ($\text{GNN}_1$ and $\text{MLP}_1$) and $f_{pred}$ ($\text{GNN}_2$ and $\text{MLP}_2$) for a fixed number of epochs $T_{sep}$ and $T_{pred}$, respectively.
The loss functions for training \method are
\begin{align}
    \label{eq:loss_pred}
    &\mathcal{L}_{pred} = \mathcal{L}_{{rem}} + \alpha \cdot \mathcal{L}_{{{rep}}}, \\
    \label{eq:loss_sep}
    &\mathcal{L}_{sep} = \mathcal{L}_{{rem}} + \alpha \cdot \mathcal{L}_{{{rep}}} + \beta \cdot  \mathcal{L}_{reg},
\end{align}
where $\mathcal{L}_{pred}$ in \cref{eq:loss_pred} and $\mathcal{L}_{sep}$ in \cref{eq:loss_sep} are used to train $f_{sep}$ ($\text{GNN}_1$ and $\text{MLP}_1$) and $f_{pred}$ ($\text{GNN}_2$ and $\text{MLP}_2$), respectively. $\alpha$ and $\beta$ are hyperparameters that control the weights of $\mathcal{L}_{rep}$ and $\mathcal{L}_{reg}$, respectively. During inference, $\hat{y}_{i}^{(r)}$ is used as the final predicted property of input graph $g_i$.

\section{Experiments}
\label{sec:experiments}
We conduct experiments to answer the following questions:
\begin{compactitem}
\item \textbf{Q1)} Effectiveness: Does the proposed \method make more accurate prediction on molecule and polymer properties than existing graph classification/regression methods?
\item \textbf{Q2)} Ablation study: Do the environment-based augmentations make positive effect on the performance?
\item \textbf{Q3)} Case study: Based on domain expertise, are the polymer rationale examples identified by \method representative? 
\item \textbf{Q4)} Efficiency: Does the \emph{latent space-based design} for augmentations perform faster than explicit graph decoding and encoding? Can we empirically analyze the complexity?
\item \textbf{Q5)} Sensitivity analysis: Is the performance of \method sensitive to hyperparameters such as $\alpha$, $\beta$, and $\text{AGG}(\cdot)$?
\end{compactitem}

\vspace{-0.05in}
\subsection{Experimental Settings}

\subsubsection{Datasets} 
We conduct experiments on \textbf{four} polymer datasets and \textbf{seven} molecule datasets. 
The statistics of the datasets are given in \cref{tab:dataset_stat}, such as number of graphs and average size of graphs. The four datasets \glassTemp, \meltTemp, \density, and \oxygen are used to predict different properties of polymers such as \emph{glass transition temperature} ($^\circ$C), \emph{polymer density} g/cm$^3$, \emph{melting temperature} ($^\circ$C), and \emph{oxygen permeability} (Barrer). For all the polymer datasets, we randomly split by 60\%/10\%/30\% for training, validation, and test. 
Besides polymer datasets, we use seven molecule datasets from the graph property prediction task on Open Graph Benchmark or known as OGBG. For all molecule datasets, we use the scaffold splitting procedure as OGBG adopted \cite{hu2020open}. It attempts to separate structurally different molecules into different subsets, which provides a more realistic estimate of model performance in experiments \cite{wu2018moleculenet}. Dataset descriptions with details are presented in the Appendix~\ref{sec:dataset_details}. 

\vspace{-0.03in}
\subsubsection{Evaluation Metrics} On the polymer datasets, we perform the tasks of graph regression. We use the coefficient of determination (\regreRSquare) and Root Mean Square Error (\regreRMSE) as evaluation metrics according to previous works~\cite{ma2020pi1m,hu2020open}.
On the molecule datasets, we perform the tasks of graph binary classification using the Area under the ROC curve (\classifyAUC) as the metric.
To evaluate model efficiency, we use the computational time per training batch (in seconds).

\vspace{-0.03in}
\subsubsection{Baseline Methods}
There are three categories of related methods that we can compare \method with. The first category is \emph{graph pooling} methods that aim at finding (soft) cluster assignment of nodes towards aggregated representations of graph. They are \unets~\cite{gao2021graph} and \selfattn~\cite{lee2019self}.
The second category improves the \emph{optimization and generalization} of learned representations. They include \stablegnn~\cite{fan2021generalizing}, \oodgnn~\cite{li2021ood}, and \irm~\cite{arjovsky2019invariant}.
The third is \dir for \emph{graph rationale identification} that was proposed in a very recent work by \citet{wu2022discovering}.
To investigate the effect of \emph{environment replacement augmentation} (denoted by \textsc{RepAug} as a module that may be used or not in the methods), we implement two method variants: (1) \dirplusaug: We add environment-replaced augmentation to \dir \cite{wu2022discovering} to identify rationales, however, it has to explicitly decode and encode the rationales; (2) \methodnoaug: We disable the environment replacement augmentation and use only the environment removal augmentation, i.e., rationale subgraphs in \method. 
In the experiments, we study two types of GNN models (\gcn~\cite{kipf2017semi} and \gin~\cite{xu2018how}) as graph encoders for all the methods. Please refer to Appendix~\ref{sec:details} for details of implementation.

\subsection{Results on Effectiveness (Q1)}\label{sec:q1_effectivene}
\cref{tab:result_plym} presents the results on polymer property regression with \regreRSquare and \regreRMSE metrics.
\cref{tab:res_ogbg_mol} presents the results on molecule property classification using \classifyAUC.
Underlined are for the best baseline(s).
The best baseline is \oodgnn for its elimination of the statistical dependence between property-relevant graph representation and property-irrelevant graph representation. The first graph rationalization method \dir was evaluated on synthetic data \cite{wu2022discovering}; unfortunately, it performs poorly on real polymer and molecule datasets because it selects edges to create rationale subgraphs and thus loses the original contextual information of atoms in the the rationale representations. Compared to them, our \method with either \gcn or \gin consistently achieves the best performance on all the polymer and molecule datasets. On the \density dataset, \method with \gcn improves \regreRSquare over \oodgnn relatively by +3.91\%. On \meltTemp, \method with \gin produces $1.56\times$ \regreRSquare over \dir.

\subsection{Ablation Study on \method (Q2)}\label{sec:q2_abalation}
\cref{tab:result_plym,tab:res_ogbg_mol} have presented the results of \dirplusaug and \methodnoaug. \dirplusaug is a variant of baseline method \dir by enabling \emph{environment replacement augmentations} for training. \methodnoaug is a variant of our \method that disables the replacement augmentations and uses \emph{environment removal} only for training.
Clearly, \dirplusaug outperforms \dir, showing positive effect of the replacement augmentations.
And the performance of \methodnoaug is not satisfactory. Environment replacement augmentations are effective for training graph rationalization methods.

\begin{figure}[t]
    \centering
    \includegraphics[width=0.96\linewidth]{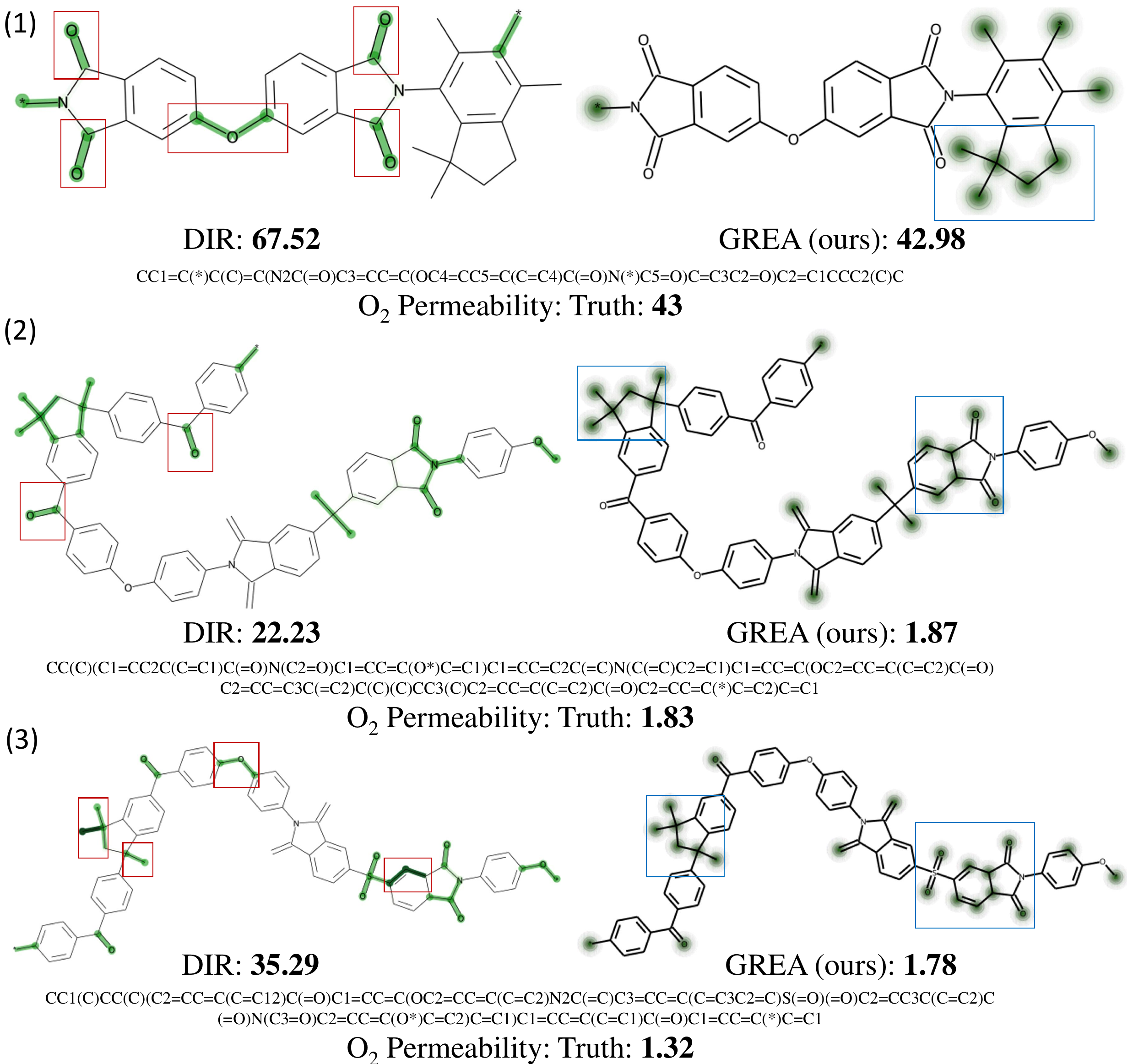}
    \vspace{-0.1in}
    \caption{Three polymer examples in O$_2$Perm test set to compare graph rationales and property predictions by \dir \cite{wu2022discovering} and our \method. \dir selects \emph{edges} to decode rationale subgraphs. Our \method estimates the probability of \emph{nodes} being classified into rationales in latent space. The red boxes indicate incoherent edges that \dir selects. The blue boxes indicate coherent node sets that contribute to accurate predictions on oxygen permeability of polymer membrane.}
    \label{fig:case_study}
    \vspace{-0.15in}
\end{figure}
\subsection{Case Study on Polymer Data (Q3)}\label{sec:q3_casestudy}
\begin{figure}[t]
    \centering
    \subfigure[Our \method runs much faster than \dir when batch size (\# graphs) increases.]
    {\includegraphics[width=0.49\linewidth]{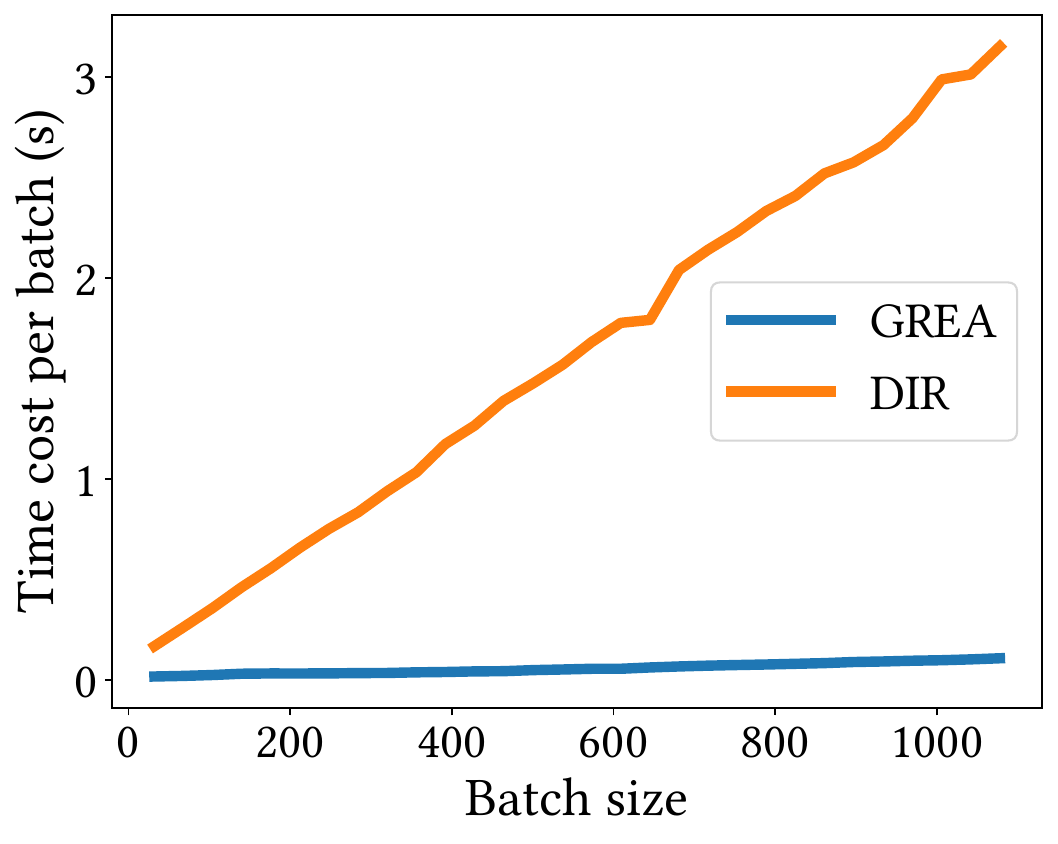}\label{fig:time_cost}}
    \hfill
    \subfigure[\method spends comparable amount of training time to deliver the highest \classifyAUC.]
    {\includegraphics[width=0.49\linewidth]{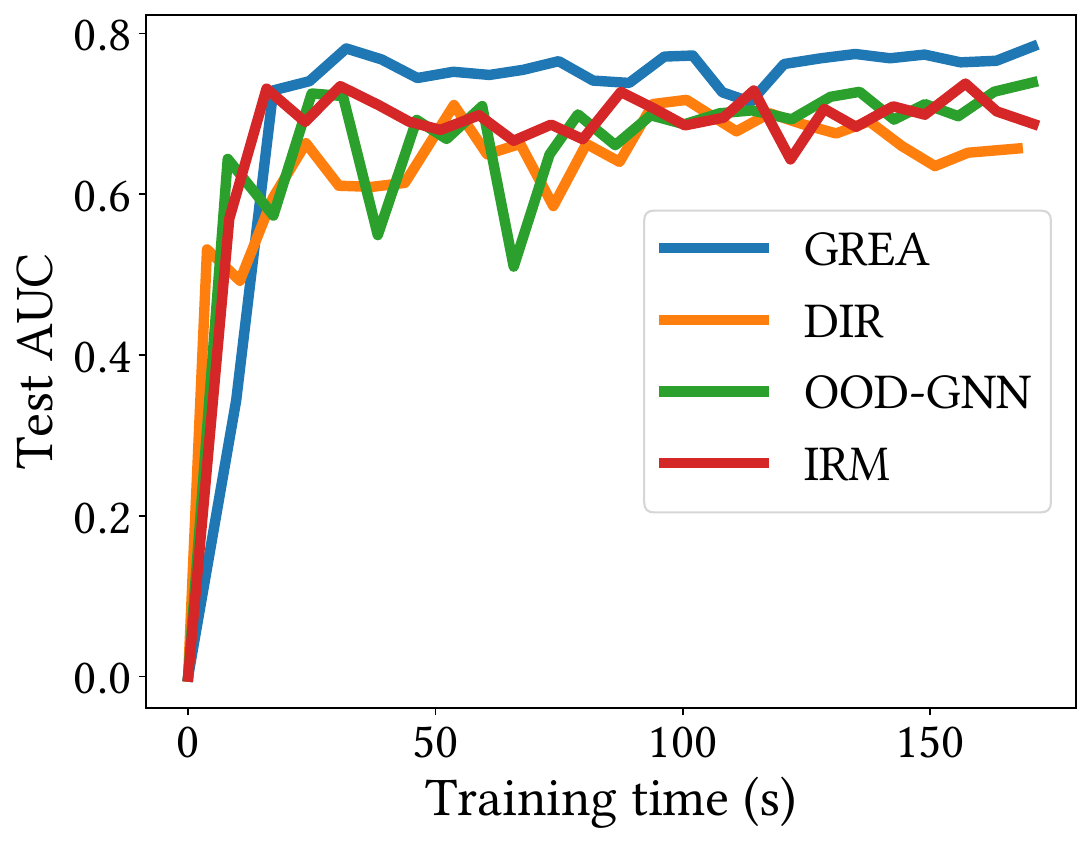}\label{fig:test_auc}}
    \vspace{-0.05in}
    \caption{Efficiency analysis on the \hiv dataset.}
    \label{fig:efficiency}
\end{figure}
\begin{figure}[t]
    \centering
    \subfigure[\density]
    {\includegraphics[width=0.49\linewidth]{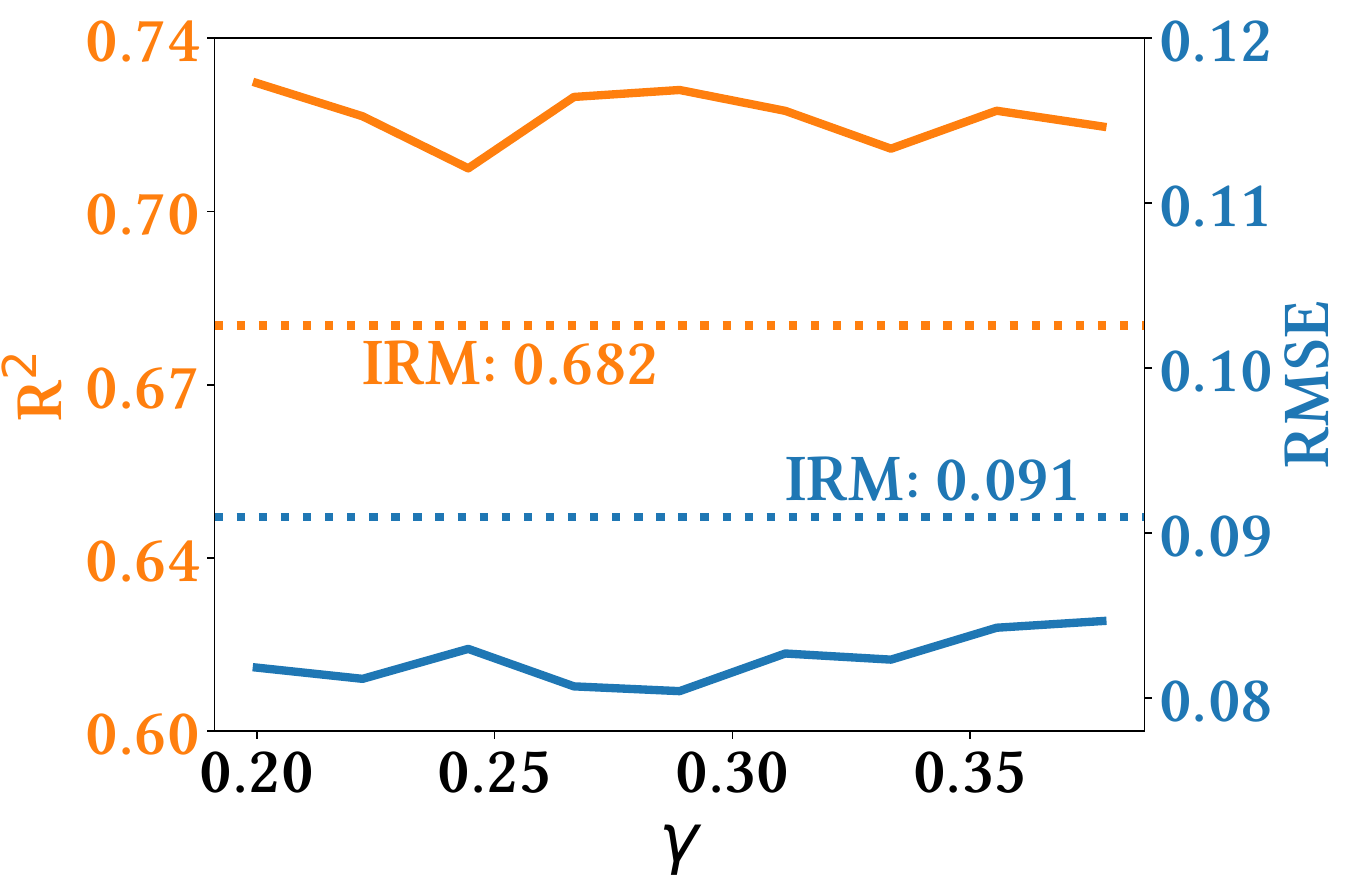}\label{fig:sensitivity_gamma_density}}
    \hfill
    \subfigure[\oxygen]
    {\includegraphics[width=0.49\linewidth]{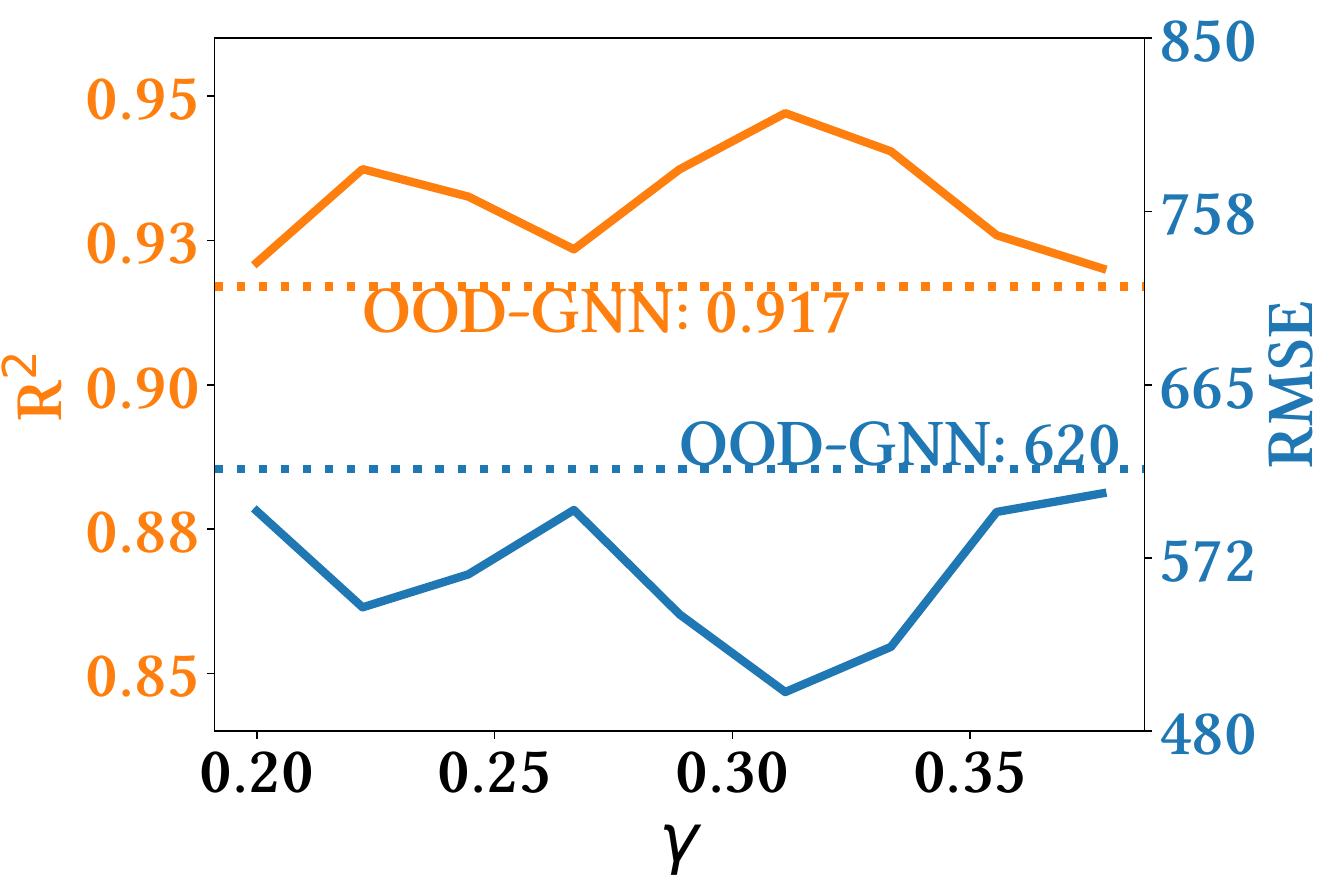}\label{fig:sensitivity_gamma_o2perm}}
    \vspace{-0.1in}
    \caption{On two polymer datasets, the performance of \method is \emph{not} sensitive to rationale size $\gamma$ with wide ranges for tuning.}
    \label{fig:sensitivity_gamma}
    \vspace{-0.05in}
\end{figure}

\begin{table}[t]
\caption{Effect of $\text{AGG}\big(\mathbf{h}_i^{(r)}, \mathbf{h}_j^{(e)}\big)$ in \cref{eq:rat_env_agg}. We use Sum Pooling by default because it generally performs the best.}
\label{tab:sensitivity_agg}
\vspace{-0.15in}
\centering
\Scale[0.86]{\begin{tabular}{l||c|c||c}
    \toprule
    & \meltTemp (\regreRSquare) & \oxygen (\regreRSquare) & \hiv (\classifyAUC) \\
    \midrule
    Sum Pooling
    & \textbf{0.7362}$\pm$0.0115 & \textbf{0.9304}$\pm$0.0202 & \textbf{0.7932}$\pm$0.0092 \\
    Mean Pooling
    & 0.7328$\pm$0.0068 & 0.9288$\pm$0.0331 & 0.7810$\pm$0.0117 \\
    Max Pooling
    & 0.7164$\pm$0.0094 & 0.8984$\pm$0.0494 & 0.7809$\pm$0.0137 \\
    Concatenation
    & 0.7145$\pm$0.0127 & 0.9240$\pm$0.0143 & 0.7771$\pm$0.0096 \\  
    \bottomrule
\end{tabular}}
\end{table}

Given test polymer examples in the \oxygen dataset, we visualize and compare the rationale subgraphs that are identified by from \dir \cite{wu2022discovering} and our \method in \cref{fig:case_study}. We have three observations.

First, the rationales identified by \method have more \emph{coherent structures of atom nodes} than those identified by \dir.
The red boxes show that quite a few edges in the rationales by \dir are far separated.
This is because \dir explicitly decodes the subgraphs by selecting edges.
Our \method estimates the probability of \emph{nodes} being included in the rationales and uses the \emph{contextualized representations} of atoms in the input graphs to create the representations of rationales.
So the rationales have coherent structures of nodes.

Second, the rationales from \method are \emph{more interpretable and beneficial} than the ones from \dir, based on domain expertise in polymer science.
Take a look at the first polymer example in Figure~\ref{fig:case_study}.
The rationale from \method includes non-aromatic rings and methyl groups.
The former group allows larger free volume elements and lower densities (i.e., enlarge microporousity) in the polymer's repeating units, which positively contributes to the gas permeability \cite{sanders2013energy,yang2021discovery}.
The latter group is hydrophobic and contributes to steric frustration between polymer chains \cite{yang2021discovery}, inducing a positive correlation to the permeability.
On the other hand, the rationale from \dir would make property predictor overestimate the oxygen permeability, because it suggests that the double-bonded oxygens, ethers, and nitrogen atoms are positively correlated with the property.
However, it conflicts with observations and conclusions from chemical experiments in previous literature \cite{yang2021discovery} where researchers argue that the double-bonded oxygens, ethers, and nitrogen atoms are negatively correlated with gas permeability.
For the second and third examples, \dir also predicts through double-bonded oxygens, ethers, and nitrogen atoms, and it overestimates the permeability.
Our \method realizes and employs the true relationship between the functional groups and property and successfully suppresses the representations of non-aromatic rings and methyl groups in the prediction.
\method intrinsically discovers correct relationships between rationale subgraphs and the property.

Third, the rationales from \method are \emph{commonly observed across different polymers}.
We expect rationales to have universal indication on the polymer properties.
The rationales identified in the second and third examples both have the fused heterocyclic rings (at the right end of the monomers and highlighted by blue boxes).

\subsection{Results on Efficiency (Q4)}\label{sec:q4_efficiency}
We conduct efficiency analysis using the \hiv dataset without losing the generality. Results are presented in Figure~\ref{fig:efficiency}. When batch size increases, in other words, when a batch has more and more graphs, the time cost per batch of \dir increases significantly; our proposed \method spends much less time than \dir. Empirically we show that our \method is more efficient than \dir. This is because \method does not explicitly decode or encode the subgraphs but directly creates their representations in latent space.
\cref{fig:test_auc} shows that compared to three most competitive baselines, \method delivers the highest \classifyAUC by learning augmented examples, while spending comparable amount of time. 

\subsection{Sensitivity Analysis (Q5)}\label{sec:q5_sensitivity}
\begin{figure}[t]
    \centering
    \vspace{-0.15in}
    \subfigure[\glassTemp]
    {\includegraphics[width=0.49\linewidth]{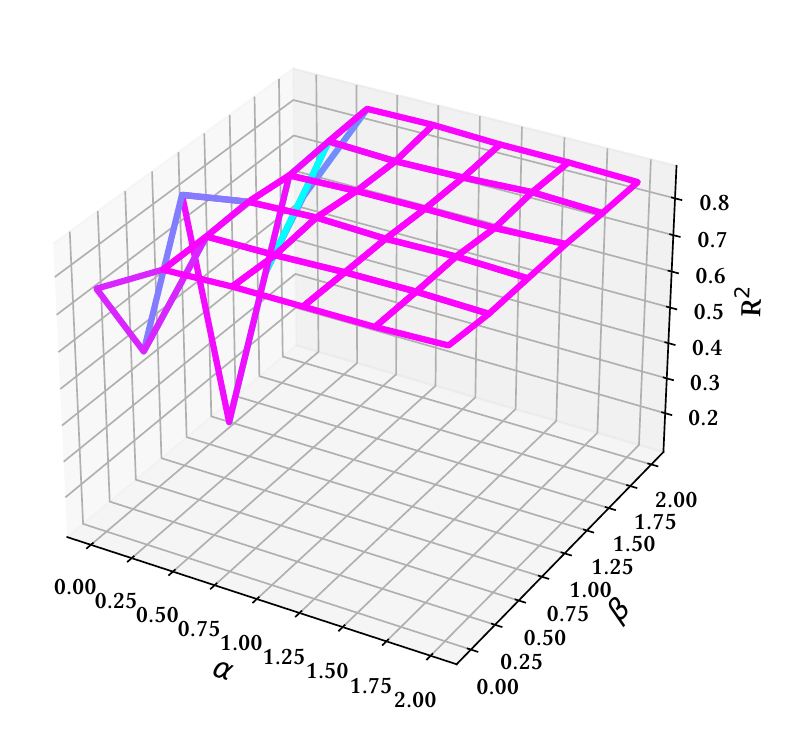}\label{fig:loss_glassT}}
    \hfill
    \subfigure[\meltTemp]
    {\includegraphics[width=0.49\linewidth]{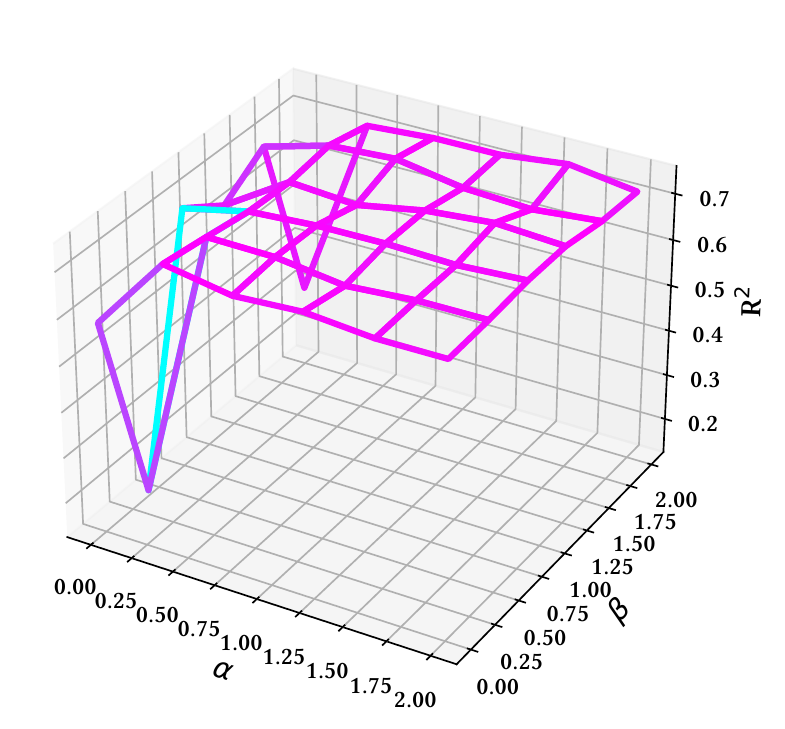}\label{fig:loss_meltT}}
    \\ \vspace{-0.15in}
    \subfigure[\density]
    {\includegraphics[width=0.49\linewidth]{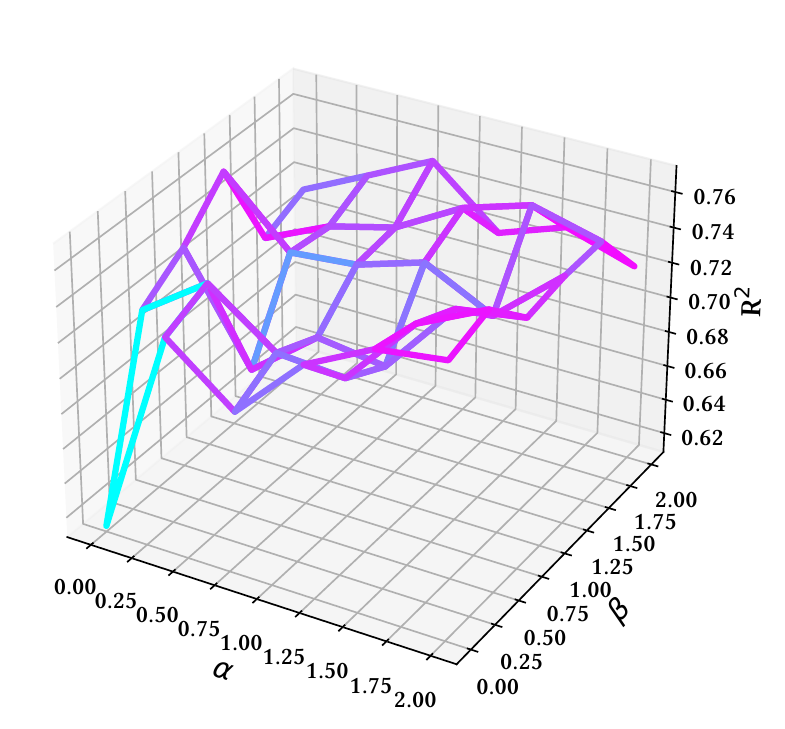}\label{fig:loss_density}}
    \hfill
    \subfigure[\oxygen]
    {\includegraphics[width=0.49\linewidth]{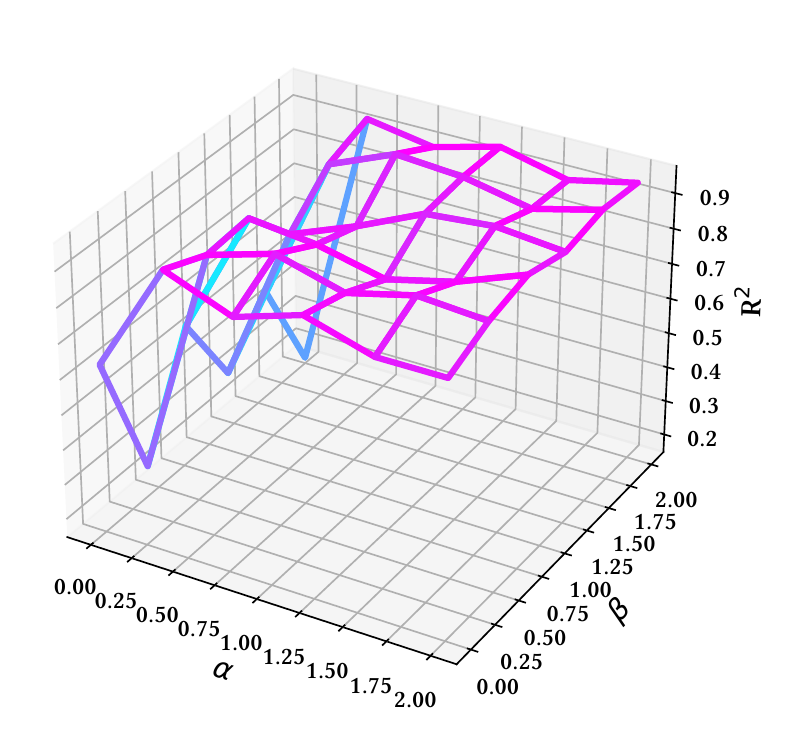}\label{fig:loss_oxygen}}
    \vspace{-0.15in}
    \caption{On four polymer datasets, the performance of \method (in \regreRSquare) is \emph{not} sensitive to hyperparameters $\alpha$ and $\beta$ in \cref{eq:loss_sep}.}
    \label{fig:loss_sensitivity}
    \vspace{-0.2in}
\end{figure}

Without losing the generality, we conduct three series of sensitivity analyses. First, \cref{fig:loss_sensitivity} shows that on four polymer datasets, the performance of \method in terms of \regreRSquare is insensitive to the hyperparameters $\alpha$ and $\beta$ in \cref{eq:loss_sep}.
Second, \cref{fig:sensitivity_gamma} shows that the performance is insensitive to rationale size $\gamma$ in \cref{eq:loss_reg}.
Third, on two polymer datasets and one of the most popular molecule datasets, \cref{tab:sensitivity_agg} compares the effects of different choices of $\text{AGG}(\cdot)$ function that aggregates the representations of rationale and environment subgraphs. Sum pooling is generally the best choice.

\section{Conclusions}
\label{sec:conclusion}
In this work, we made the first attempt to improve graph rationale identification using data augmentations, including environment replacement, for accurate and interpretable graph property prediction. We proposed an efficient framework that performs rationale-environment separation and representation learning on real and augmented examples in one latent space. Experiments on molecule and polymer datasets demonstrated its effectiveness and efficiency.

\begin{acks}
This research was supported in part by NSF Grants IIS-1849816, IIS-2142827, IIS-2146761, and CBET-2102592.
\end{acks}

\balance
\bibliographystyle{ACM-Reference-Format}
\bibliography{ref-short}

\clearpage
\appendix
\section{Dataset Details}
\label{sec:dataset_details}
\vspace{-0.05in}
\paragraph{Polymer datasets} The four datasets \glassTemp, \meltTemp, \density, and \oxygen are used to predict different properties of polymers such as \emph{glass transition temperature} ($^\circ$C), \emph{polymer density} g/cm$^3$, \emph{melting temperature} ($^\circ$C), and \emph{oxygen permeability} (Barrer). \glassTemp, \meltTemp, and \density are collected from PolyInfo, which is the largest web-based polymer database \cite{otsuka2011polyinfo}. The \oxygen dataset is created from the Membrane Society of Australasia portal,
consisting of a variety of gas permeability data \cite{thornton2012polymer}. However, the limited size (i.e., 595 polymers) brings great challenges to rationale identification and property prediction. Since a polymer is built from repeated monomer units, researchers use monomers as polymer graphs to predict properties. Different from molecular graphs, the monomer graphs have two special nodes (see ``$*$'' in the molecular structures in Figure~\ref{fig:idea}), indicating the polymerization points of monomers \cite{ma2020pi1m}. For all the polymer datasets, we randomly split by 60\%/10\%/30\% for training, validation, and test.

\vspace{-0.05in}
\paragraph{Molecule datasets} Besides polymer datasets, we use seven molecule datasets from the graph property prediction task on Open Graph Benchmark or known as OGBG. They were originally collected by MoleculeNet \cite{wu2018moleculenet} and used to predict the properties of molecules, including (1) inhibition to HIV virus replication in \hiv, (2) toxicological properties of 617 types in \toxcast, (3) toxicity measurements such as nuclear receptors and stress response in \toxt, (4) blood–brain barrier permeability in \bbbp, (5) inhibition to human $\beta$-secretase 1 in \bace, (6) FDA approval status or failed clinical trial in \clintox, and (7) having drug side effects of 27 system organ classes in \sider. For all molecule datasets, we use the scaffold splitting procedure as OGBG adopted \cite{hu2020open}. It attempts to separate structurally different molecules into different subsets, which provides a more realistic estimate of model performance in experiments \cite{wu2018moleculenet}.

\section{Implementation Details}
\label{sec:details}

All the experiments in this work are conducted on an Linux server with Intel Xeon Gold 6130 Processor (16 Cores @2.1Ghz), 96 GB of RAM, and a single RTX 2080Ti card (11 GB of RAM). Our method is implemented with \verb+Python 3.9.9+ and \verb+PyTorch 1.10.1+. We manually tune the hyperparameters over the following ranges:
\begin{compactitem}
\item $\gamma \in \{0.05, 0.1, 0.15, \dots, 0.75, 0.8\}$,
\item $T_{sep} \in \{1,2\}$,
\item $T_{pred} \in \{2,3\}$,
\item Learning rate $\in \{0.001, 0.005, 0.01\}$,
\item Batch size $\in \{32, 128, 256, 512\}$,
\item Representation dimensions $d_1$, $d_2 \in \{64, 128, 300\}$,
\item Number of GNN$_1$ layer $L_1=\{2\}$,
\item Number of GNN$_2$ layers $L_2 \in \{2,3,4,5\}$.
\end{compactitem}
We use sum pooling as the default $\text{AGG}(\cdot)$ in \method for the experiments in \cref{tab:result_plym,tab:res_ogbg_mol}. We set \gin as the default encoder for all ablation studies, case studies, and efficiency analysis.
We employ the virtual node trick~\cite{hu2020open} for all methods on the \hiv, \toxt, \bbbp, and all polymer datasets. For \density, we train and evaluate the models using the logarithm of the property~\cite{ma2020pi1m}.
We report the mean and standard deviation of the test performance over 10 runs with different random initialization of the parameters. 

Our code and data are available on the GitHub\footnote{\url{https://github.com/liugangcode/GREA}}. To implement the baseline methods, we use the official code package\footnote{\url{https://github.com/Wuyxin/DIR-GNN}} from the authors for \dir~\citep{wu2022discovering}. For \unets~\cite{gao2021graph} and \selfattn~\cite{lee2019self}, we use the public implementation provided by the \verb+PyG+\footnote{\url{https://github.com/pyg-team/pytorch_geometric}} package. For \irm~\citep{arjovsky2019invariant}, we implement it's graph version based on its official repository.\footnote{\url{https://github.com/facebookresearch/InvariantRiskMinimization}}
As source codes of \oodgnn~\cite{li2021ood} and \stablegnn~\cite{fan2021generalizing} are not publically available, we implement then with the official code package of \textsc{StableNet}\footnote{\url{https://github.com/xxgege/StableNet}} and the \verb+PyG+ package.

\end{document}